\documentclass[sigconf]{acmart}
\AtBeginDocument{%
  \providecommand\BibTeX{{%
    \normalfont B\kern-0.5em{\scshape i\kern-0.25em b}\kern-0.8em\TeX}}}

\setcopyright{acmcopyright}
\copyrightyear{2024}
\acmYear{2024}

\acmConference[HRI'24]{HRI 101 Workshop}{March 11--15, 2024}{Boulder, CO}
\usepackage{tikz}
\usepackage{enumitem}
\setlist[itemize]{leftmargin=5mm}

\begin{document}

\title{ROB 204: Introduction to Human-Robot Systems\\ at the University of Michigan, Ann Arbor}

\author{Leia Stirling}
\authornotemark[1]
\email{leias@umich.edu}
\orcid{0000-0002-0119-1617}
\affiliation{%
  \institution{University of Michigan}
  \city{Ann Arbor}
  \state{MI}
  \country{USA}
  \postcode{48109}
}

\author{Joseph Montgomery}
\email{montgjoa@umich.edu}
\affiliation{%
  \institution{University of Michigan}
  \city{Ann Arbor}
  \state{MI}
  \country{USA}
  \postcode{48109}
}

\author{Mark Draelos}
\email{mdraelos@umich.edu }
\affiliation{%
  \institution{University of Michigan}
  \city{Ann Arbor}
  \state{MI}
  \country{USA}
  \postcode{48109}
}

\author{Christoforos Mavrogiannis}
\email{cmavro@umich.edu}
\affiliation{%
  \institution{University of Michigan}
  \city{Ann Arbor}
  \state{MI}
  \country{USA}
  \postcode{48109}
}

\author{Lionel P. Robert Jr.}
\email{lprobert@umich.edu }
\affiliation{%
  \institution{University of Michigan}
  \city{Ann Arbor}
  \state{MI}
  \country{USA}
  \postcode{48109}
}

\author{Odest Chadwicke Jenkins}
\email{ocj@umich.edu}
\affiliation{%
  \institution{University of Michigan}
  \city{Ann Arbor}
  \state{MI}
  \country{USA}
  \postcode{48109}
}

\renewcommand{\shortauthors}{Stirling et al.}

\begin{abstract}
The University of Michigan Robotics program focuses on the study of embodied intelligence that must sense, reason, act, and work with people to improve quality of life and productivity equitably across society. ROB 204, part of the core curriculum towards the undergraduate degree in Robotics, introduces students to topics that enable conceptually designing a robotic system to address users' needs from a sociotechnical context. Students are introduced to human-robot interaction (HRI) concepts and the process for socially-engaged design with a Learn-Reinforce-Integrate approach. In this paper, we discuss the course topics and our teaching methodology, and provide recommendations for delivering this material. Overall, students leave the course with a new understanding and appreciation for how human capabilities can inform requirements for a robotics system, how humans can interact with a robot, and how to assess the usability of robotic systems.
\end{abstract}


\received{21 January 2024}
\received[accepted]{6 February 2024}

\maketitle

\section{Introduction}
\begin{figure*}[tb] 
\centering
\includegraphics[width=0.85\linewidth]{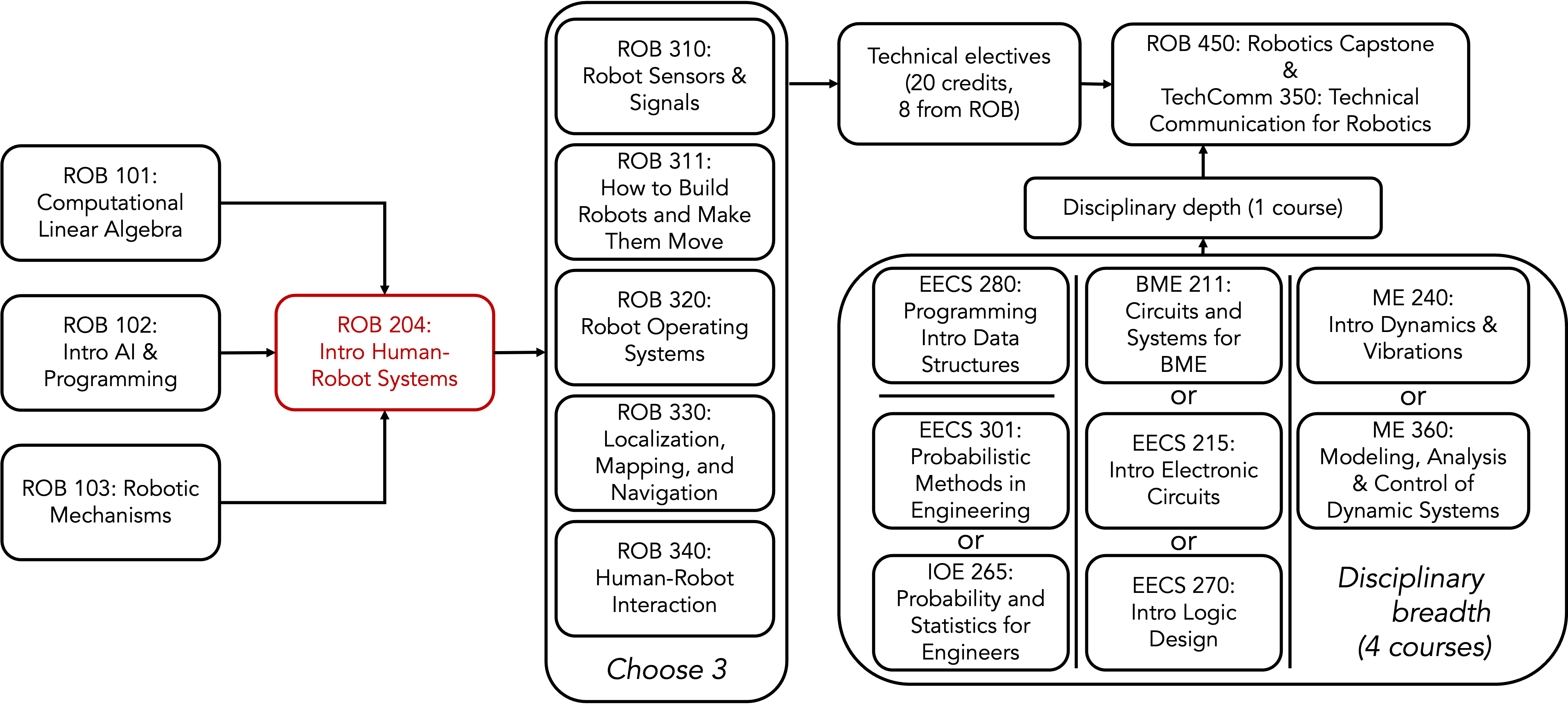}
\caption{The University of Michigan Robotics curriculum includes different pathways for students to gain breadth of knowledge, while also enabling selection of courses to build knowledge depth.}
\label{Fig:ROBCurriculum}
\end{figure*}

The University of Michigan Robotics undergraduate program focuses on the study of embodied intelligence that must sense, reason, act, and work with people to improve quality of life and productivity equitably across society. ROB 204: Introduction to Human-Robot Systems serves as the gateway into the intermediate level (Fig.~\ref{Fig:ROBCurriculum}) for robotics majors and introduces students to topics that enable conceptually designing a robotic system to address a user need from a sociotechnical context. Students are introduced to human-robot interaction concepts and the process for socially-engaged design \cite{CSED} with a Learn-Reinforce-Integrate approach. Students begin by learning new concepts and skills in interactive lectures, reinforce the technical concepts in labs, and then integrate the concepts through comprehensive assessments. 

The course combines technical and technical communication elements to support students building and communicating their understanding. Students are provided with frameworks to address the following technical questions:
\begin{itemize}
\item What process should we use to ideate robotic system designs?
\item How can a person interact with a robotic system?
\item How can characteristics of the users influence the requirements for a robotic system?
\item How can we assess our design ideas with end users?
\end{itemize}
and the following technical communication questions:
\begin{itemize}
\item How can we use written, visual, and oral communication to engage with both non-engineers and engineers to support the design process and integration of robots within society?
\item How can we read and use scholarly articles to support the design process?
\end{itemize}

In Section \ref{Sec:Syllabus Content}, we provide additional details on the lecture topics. In Section \ref{Sec:Teaching methodologies}, we discuss our approach to delivering the content and assessing student learning. Finally, in Section \ref{Sec:Discussion}, we provide our thoughts on the material integrated and the interconnectivity with other courses in the curriculum that support the education of a Michigan roboticist. 

\section{Technical Topics and Schedule}
\label{Sec:Syllabus Content}

ROB 204 includes weekly lectures and labs to introduce content and practice the material (Table \ref{Tab:schedule}). The first half of the term introduces students to concepts in human perception, human cognition, and human-robot communication. The second half of the term introduces a socially-engaged design process, with a focus on generating problem statements and system requirements and assessing system usability.

The technical content of the course is organized along the following themes, with example learning objectives provided:\newline

\pagebreak 
\noindent\textbf{Human perception and user interfaces}: Learn how principles of human perception inform robotic system design.
\begin{itemize}
    \item Describe how the human sensory system processes light, sound, and physical contact.
    \item Apply concepts of the human sensory system to robotic design requirements.
    \item Design a graphical user interface using principles of effective visual displays (e.g., Gestalt principles, data- and goal-driven processing, influence of peripheral/foveal vision).
    \item Describe how true, false, missed, and nuisance auditory alarms can affect safety.
\end{itemize}


\noindent\textbf{Human cognition}: Learn how concepts from human decision making inform robot design requirements. 
\begin{itemize}
    \item Describe the human information processing model.
    \item Compare/contrast normative and descriptive decision making.
    \item Give examples of how automation level can vary between the different information processing stages.
    \item Critique the design of a robotic application to support or enhance situation awareness, trust, and transparency.
\end{itemize}

\noindent\textbf{Human in the loop communication}: Learn how human intelligence, decision-making, and oversight within communication systems affect human-robot interaction.
\begin{itemize}
    \item Identify non-verbal and verbal interactions between a human and robot.
    \item Describe how robot behavior can be informed by human inputs.
    \item Critique the selection of a control input for a robot given a use case and scenario.
\end{itemize}
 
\noindent\textbf{Ethics:} Learn about guidelines, principles, and frameworks for the ethical design, development, deployment, and use of robotic and autonomous systems. 
\begin{itemize}
    \item Identify and describe ethical considerations for robotic systems.
    \item Discuss ethical viewpoints in the context of a case study.
    \item Apply the intention-action-consequences framework to define intentions and to avoid negative consequences.
    \item Present design concepts that reframe and limit ethical dilemmas.
\end{itemize}

\noindent\textbf{Problem statements, needs statements, requirements}: Understand how the design process is informed by user needs and requirements.
\begin{itemize}
    \item Develop and critique problem statements, needs statements, and system requirements. 
    \item Critique an operational design domain for a given scenario.
    \item Identify sources of risk and opportunities for risk reduction.
\end{itemize}

\noindent\textbf{Socially engaged design:} Understand how to create solutions that address user needs by collaborating with stakeholders.

\begin{itemize}
    \item Describe how identity and power can influence community engagement for human-centered design.
    \item Apply active listening techniques to support stakeholder engagement.
    \item Develop and carry out an interview protocol that meets best practices.
\end{itemize}

\noindent\textbf{Usability studies:} Learn about usability concepts and methods for operationalizing and assessing usability.
\begin{itemize}
    \item Describe learnability, efficiency, memorability, fluency, error, satisfaction, and trust.
    \item Compare and contrast methods of usability testing.
    \item Provide examples of operationalizing a usability concept at different levels of measurement.
    \item Design a usability study to assess a robotic system.
\end{itemize}

\begin{table*}[t]
\centering
\caption{Schedule of technical content and lab activity for ROB 204}
\begin{tabular}{l|l|l}
 \text{Week}& Content & Lab Activity  \\
 \hline
 1 -- 2  & Human perception and user interfaces & User interface design and evaluation \\
 3 -- 4  & Human cognition (mental models; situation awareness; trust) & Mental model building and assessment   \\
 5 -- 6  & Human in the loop communication & Levels of control for a mobile robot  \\
 7 -- 8  & Ethics; socially engaged design & Stakeholder interviews  \\
 9      & Problem statements, needs statements, and requirements &  Problem statement generation \\
 10 -- 11 & Usability studies & Robot concept design, usability study design \\
 12 -- 15 & Final project work sessions and presentations & Usability study, design iteration  
\end{tabular}
\label{Tab:schedule}
\end{table*}

\section{Teaching Methodologies}
\label{Sec:Teaching methodologies}
ROB 204 is taught as a lecture with additional laboratory sessions. The lecture sessions introduce learning objectives aligned with the week's theme (Table \ref{Tab:schedule}). These learning objectives follow Bloom's Taxonomy \cite{anderson2001taxonomy}, introducing concepts to support remembering, understanding, and applying, with the goal of enabling students to analyze, evaluate, and create. Content is introduced through definitions, conceptual examples, and case-studies of deployed systems. Embedded in these lectures are small group discussions that allow students to engage with each other and the teaching team and to share their thoughts in a less intimidating manner than through speaking in front of the entire class, while also providing an opportunity to correct misinterpretations that may arise.

At the end of most lectures, a student team presents a journal article related to that week's theme. These articles serve two purposes: (1) they provide additional HRI examples that highlight applications, methods, and challenges that are being addressed in the research community and (2) they enable practice in reading scholarly articles and summarizing important information (why is the question important to consider, what is the specific goal of the paper, how did they approach their question, what is something that was learned, how can what was learned be applied). 

\begin{figure}[b]
    \centering
    \includegraphics[width=\linewidth]{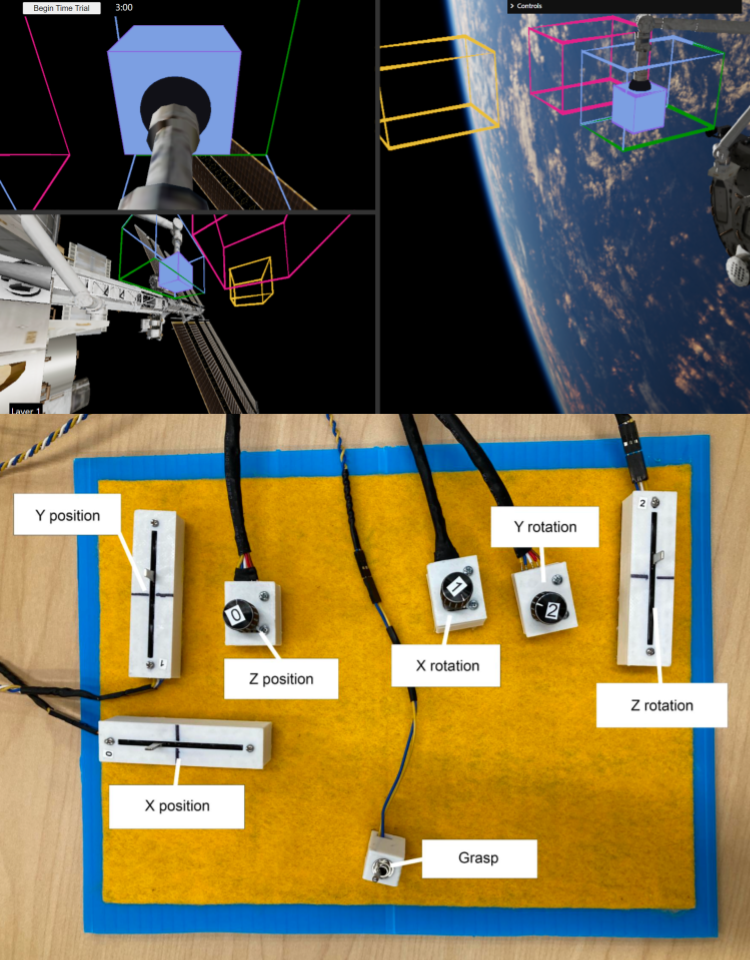}
    \caption{%
        Labs 1 and 2 teach user interface and control strategies.
        Students build their own physical interface (bottom) to control a simulated Canadarm (top).
        Exercises teach control from fixed and local coordinate frames and effect of system order (e.g., controlling position vs. velocity).
        \label{fig:simulator}
    }
\end{figure}

\begin{figure}[t]
    \centering
    \includegraphics[width=\linewidth]{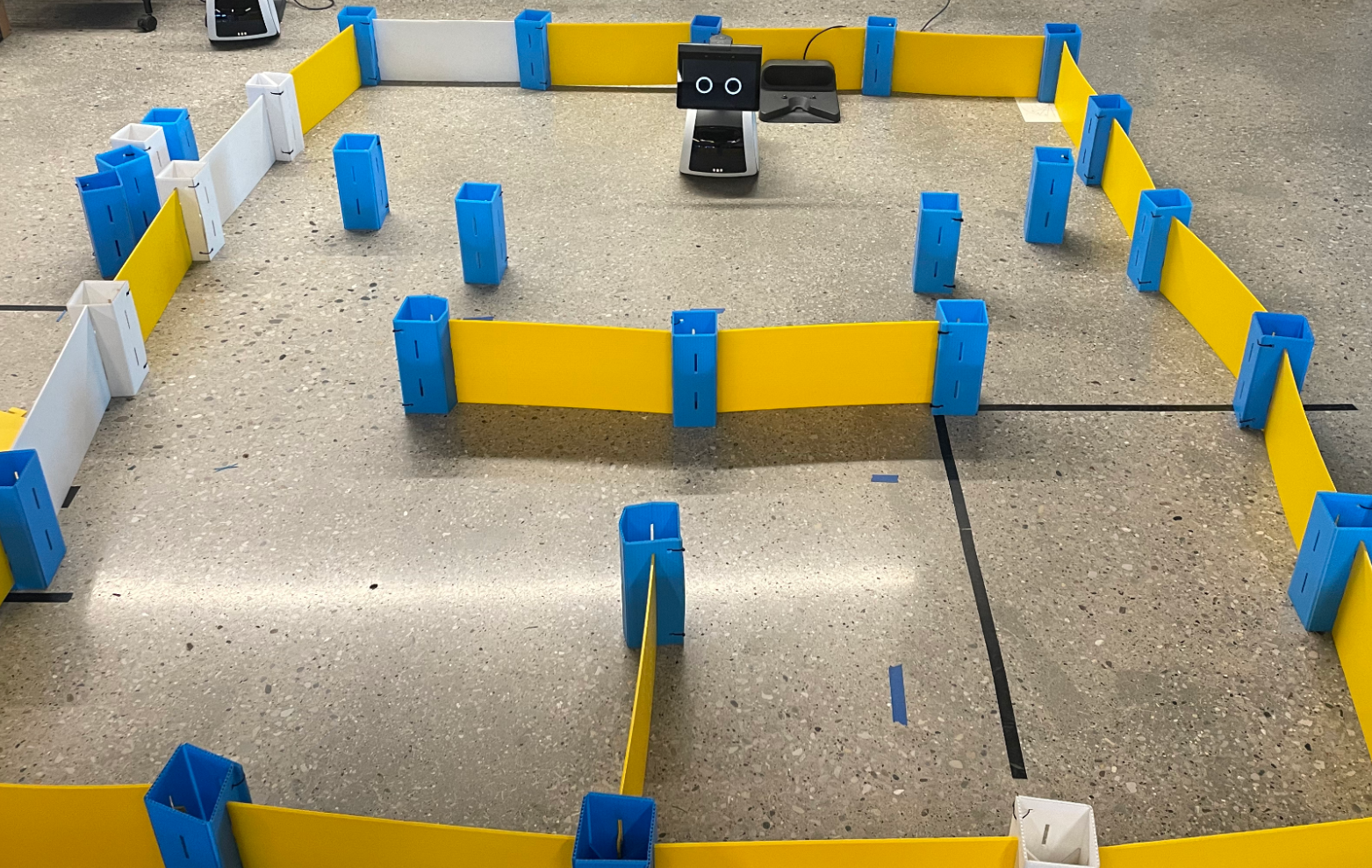}
    \caption{
        Labs 3 and 4 teach trust calibration for robots.
        Students perform a series of navigation tasks with the Amazon Astro in a maze with varied obstacles.
        Students rate their level of confidence in the robot's ability to successfully carry out the tasks.
        \label{fig:maze}
    }
\end{figure}

Lab sessions build on the lectures and small-group interactions by providing students with guided, emergent, hands-on learning experiences with robotic systems and socially-engaged design principles. During the 2023-24 academic year, the labs included designing and testing physical user interfaces to control a simulated robotic arm in a space environment (Fig.~\ref{fig:simulator}) and exploring levels of control, trust, and mental models with the Amazon Astro (Fig.~\ref{fig:maze}).

During the second half of the semester, students use the lab sessions to work towards the final project: the conceptual design and evaluation of a robot to assist a nurse. Students write interview questions and interview several nurse volunteers from University of Michigan Medicine. Based on their interview findings, students develop problem statements and system requirements and then begin to design a robot. They conduct library research to better understand the problem they've identified and see how others have addressed it. They then design and conduct a usability study on their initial design, again with the nurse volunteers, and then iterate on their design before presenting it to the class.

In addition to the labs and the final project, student learning and performance is comprehensively assessed through two reflection assignments, which are take-home assignments that take the place of a midterm and final exam. Each reflection includes short answer prompts on various technical learning objectives, along with a writing assignment that combines technical and technical communication learning objectives. During the 2023-24 academic year, the first reflection included a memo that presented test results through writing and visuals and made recommendations to a manager, and the second reflection asked students to define a research question related to a course topic and create an annotated bibliography.

Throughout the semester, students build upon the technical communication skills introduced in Eng 100: Introduction to Engineering. A few lecture sessions are used to discuss communication topics, each lab session includes communication practice, and each major assignment assesses technical communication skills alongside technical competency. Communication skills include finding and analyzing scholarly articles, designing tables and charts, writing memos, and delivering oral presentations.

\section{Discussion}
\label{Sec:Discussion}
Through the University of Michigan Robotics undergraduate curriculum, students build expertise in hardware-focused mechatronics, computing-focused autonomy, and empiricism-focused human-robot interaction \cite{jenkins2023michigan}. ROB 204 introduces students to human sensory and cognitive capabilities and to different modalities of human-robot communication. These concepts support a socially-engaged design process that includes defining system requirements, ideating robotic concepts, and assessing system usability. By offering a Robotics major rather than a single Robotics course, we can introduce HRI content early on, which students then build upon in upper-level courses.  

Many of the concepts taught in the course are abstract; the application of the theory is not an algorithm. For example, usability studies differ in design based on the goals of the robot and the method of human interaction. For many students, this type of divergent thinking is new, but is important to build early in their education. 

End-of-semester course evaluations support desired course takeaways. Students noted that ``people have different needs and wants'' than those of the designer. They were also surprised at how much could be learned through conceptual design prior to building a physical prototype: ``getting feedback from users can start much earlier in the design process and can happen more frequently than we expected.''  Students were able to connect the learning objectives on human capabilities to their design process: ``understanding specifics about how the human works made designing effective systems much easier.'' Another student commented that ``as engineers, we are tasked with the responsibility of ensuring that the future of robotics is not only innovative, but practical and safe for humanity.'' This takeaway  aligns with the University of Michigan vision for ``People-first Engineering'' and the Robotics Department's values of ``Robotics with Respect'' and ``Integrity in Action.''




Our collective experience with ROB 204 now spans four semesters with four technical instructors and four technical communication instructors. We have distilled this experience into a list of recommendations for teaching HRI at the college sophomore level.

\begin{itemize}

\item \textbf{Avoid anthropomorphizing robots.}
Early portions of our course compare and contrast the sensing, reasoning, and acting capabilities of humans with those of robots.
When students anthropomorphize robots in class discussions or lab assignments, we have observed that they incorrectly ascribed capabilities of human to robots.
In assignments related to the types of human errors, for example, students described the robot as ``forgetting'' a task, an error only properly attributed to human memory lapses.
As robots increase in sophistication and integration in society, the ability to distinguish between the possible human and robot errors is a key competency for designers.

\item \textbf{Incorporate small group discussions.}
Designing human robot systems requires a holistic evaluation of many factors, from high-level concepts (e.g., usability) to implementation details (e.g., button placement).
We have observed that students learn to think holistically through small group discussions more effectively than through traditional lectures. 
Such discussions allow students to identify their assumptions and broaden their perspective through immediate peer feedback.

\item \textbf{Reinforce content through in-class exercises.}
We have found that students do not often invest sufficient effort in understanding HRI concepts that they consider obvious (e.g., mobile robots respecting human personal space).
When assessed through lab or written assignments, students discover their inability to operationalize these concepts (e.g., failure to address human-robot spacing in designs).
We therefore recommend reinforcement of concepts that students might initially ``write off'' through immediate in-class exercises and lab experiences.

\item \textbf{Include labs with user studies.}
Many attributes of human robot systems are best understood through user observation and interviews.
In ROB 204, students engage with users at the beginning and in the middle of their design process.
Labs such as these enable students to practice the design skills we teach and challenges their assumptions by interacting with end-users.

\end{itemize}

\section{Conclusion}
All robotic systems interact with humans at some phase of their lifecycle. As we educate future roboticists, it is critical that they build robots with human-robot interactions in mind. ROB 204 was designed to introduce students to HRI from a socially-engaged design perspective, supporting students' abilities to ideate and evaluate robotic systems with a human-centered perspective.

\bibliographystyle{plain}

\end{document}